\documentclass[letterpaper]{article} 
\usepackage[]{aaai2026}  
\usepackage{times}  
\usepackage{booktabs}
\usepackage{amssymb}

\usepackage{helvet}  
\usepackage[table]{xcolor}

\usepackage{courier}  
\usepackage[hyphens]{url}  
\usepackage{graphicx} 
\usepackage{amsmath}
\usepackage{multirow} 
\usepackage{natbib}  
\usepackage{caption} 
\usepackage{algorithm}
\usepackage{algorithmic}
\definecolor{darkgreen}{rgb}{0.0, 0.5, 0.0}
\definecolor{goldenyellow}{rgb}{1.0, 0.75, 0.0}
\usepackage{newfloat}
\usepackage{listings}
\DeclareCaptionStyle{ruled}{labelfont=normalfont,labelsep=colon,strut=off} 
\floatstyle{ruled}
\newfloat{listing}{tb}{lst}{}
\floatname{listing}{Listing}
\title{Parameter-Free Fine-tuning via Redundancy Elimination \\ for Vision Foundation
Models}
\author{
    Jiahuan Long\textsuperscript{\rm 1, 3},
    Tingsong Jiang\textsuperscript{\rm 3},
    Wen Yao\textsuperscript{\rm 3}\equalcontrib,
    Yizhe Xiong\textsuperscript{\rm 2},
    Zhengqin Xu\textsuperscript{\rm 1},
    Shuai Jia\textsuperscript{\rm 1}, \\
    Hanqing Liu\textsuperscript{\rm 1},
    Chao Ma\textsuperscript{\rm 1}\equalcontrib,
}
\affiliations{
    \textsuperscript{\rm 1}Shanghai Jiao Tong University, 
    \textsuperscript{\rm 2}Tsinghua University, 
    \textsuperscript{\rm 3}Chinese Academy of Military Science, 
\\
}

\usepackage{bibentry}

\begin{document}

\maketitle

\begin{abstract}
Vision foundation models (VFMs) have demonstrated remarkable capabilities in learning universal visual representations. 
However, adapting these models to downstream tasks conventionally requires parameter updates, with even parameter-efficient fine-tuning methods necessitating the modification of thousands to millions of weights. In this paper, we investigate the redundancies in the segment anything model (SAM) and then propose a novel parameter-free fine-tuning method. 
Unlike traditional fine-tuning methods that adjust parameters, our method emphasizes  \textbf{selecting}, \textbf{reusing}, and \textbf{enhancing} pre-trained features, offering a new perspective on fine-tuning foundation models.
Specifically, we introduce a channel selection algorithm based on the model's output difference to identify redundant and effective channels. By selectively replacing the redundant channels with more effective ones, we filter out less useful features and reuse more task-irrelevant features to downstream tasks, thereby enhancing the task-specific feature representation. 
Experiments on both out-of-domain and in-domain datasets demonstrate the efficiency and effectiveness of our method in different vision tasks (e.g., image segmentation, depth estimation and image classification). 
Notably, our approach can seamlessly integrate with existing fine-tuning strategies (e.g., LoRA, Adapter), further boosting the performance of already fine-tuned models.
Moreover, since our channel selection involves only model inference, our method significantly reduces GPU memory overhead.\end{abstract}


\section{Introduction}

\begin{figure}[t]
\centering
\includegraphics[scale=0.42]{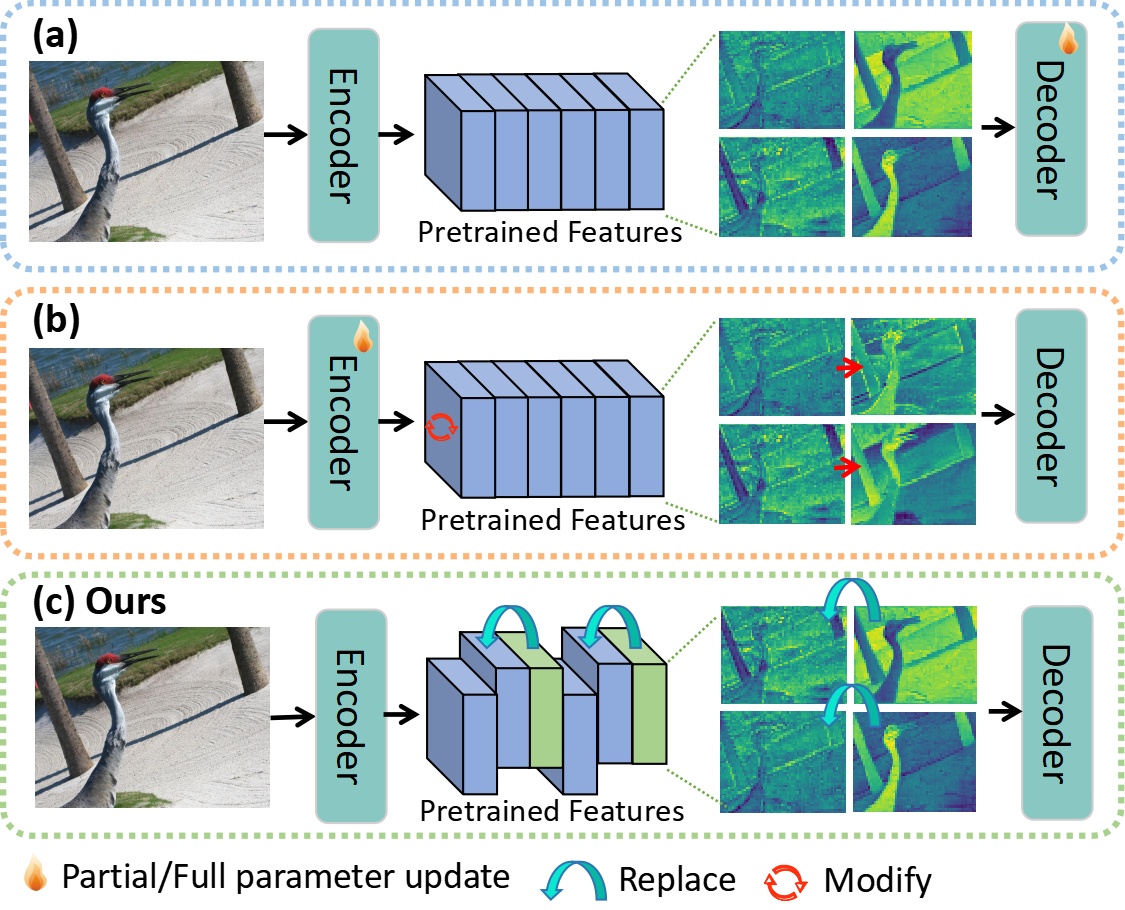}
\vspace{-0.1in}
\caption{
\textbf{A comparison between our method and other fine-tuning methods.} (a) shows a fine-tuning method that updates the decoder to align pretrained features with the downstream task; (b) illustrates a fine-tuning method that updates the encoder to modify pretrained features for downstream adaptation. In contrast, 
(c) depicts our method, which adapts pretrained features to the downstream task by replacing specific redundant channels without any parameter updates.
}
\label{fig:feature comparison}
\end{figure}

Vision foundation models (VFMs)~\cite{SAM, SAM2, CLIP, dinov2} are large-scale and generalizable models designed to adapt to various visual tasks, such as image classification, detection, and segmentation. 
Among the VFMs, Segment Anything Model (SAM)~\cite{SAM, SAM2} 
significantly advances the research of vision foundation models on various visual tasks~\cite{trackinganything, groundedsam, Inpaintanything}.
Fine-tuning SAM can further unleash its potential in new specific tasks or scenarios, such as camouflage detection~\cite{SAM-adapter-Camouflage-Shadow-more} and medical image segmentation~\cite{samed, MedSAM}, 
achieving superior performance than prior task-specific models. 

Recently, parameter-efficient fine-tuning (PEFT)~\cite{xiong2025pyra, LoRA, adapter-icml, SVDdiff, zelinSVD} methods have become mainstream approaches for adapting VFMs to downstream tasks. Compared to full-parameter fine-tuning, PEFT updates only a small subset of parameters while keeping the majority of pre-trained parameters unchanged, significantly reducing computational overhead. 
For example, SAMed~\cite{samed} customizes SAM for medical image segmentation using only 4.4\% of SAM’s parameters via low-rank adaptation~\cite{LoRA}. Another approach, Sam-PARSER~\cite{zelinSVD}, further reduces parameter usage to just 512 learnable parameters while still achieving substantial performance gains on medical datasets. As PEFT methods continue to shrink the number of trainable parameters, a radical idea arises: \textit{can a VFM be adapted to downstream tasks without modifying any model parameters?}


From the perspective of feature representation, existing works indicate that deep learning models exhibit significant feature redundancy~\cite{li2023scconv, dalvi2020analyzing}. This is largely because networks trained on large-scale datasets tend to learn universal features to generalize across different tasks, many of which may be irrelevant or even detrimental to specific downstream objectives.
In particular, vision foundation models, designed to capture universal features from large-scale training data, retain substantial task-irrelevant features when adapted to smaller downstream datasets.
To illustrate this point, we conduct a controlled experiment on the Segment Anything Model (SAM), in which we manually deactivate specific feature channels by setting their activation values to zero (i.e., effectively turning off those channels).  As shown in Table~\ref{tab:PerSegEffect}, we observed that when some channels are inactive, the overall performance on PerSeg dataset remains unchanged (channel 6) and even improves (channel 216). This indicates that some channels in SAM are redundant or noisy for downstream tasks, as removing them does not impair performance (Refer to Figure~\ref{fig:Visualization of redundant and effective features} for visualizations of these redundant channels).

\begin{table}[t]
\centering
\small
\resizebox{0.47\textwidth}{!}{
\begin{tabular}{c|ccccc}
\toprule
Channel number  & 6 & 216 & 175 & 19 & 189 \\
\midrule
\textbf{Baseline} & 50.6 & 50.6 & 50.6 & 50.6 & 50.6 \\
\textbf{Inactive (value = 0)} & \textbf{50.6} \textcolor{red}{=} & \textbf{52.7} \textcolor{red}{$\uparrow$} & 48.7 \textcolor{darkgreen}{$\downarrow$} & 48.4 \textcolor{darkgreen}{$\downarrow$} & 47.9 \textcolor{darkgreen}{$\downarrow$} \\
\bottomrule
\end{tabular}
}
\vspace{-0.05in}
\caption{\textbf{Performance comparison of SAM on the PerSeg dataset with inactive channels.} }
\vspace{-0.2in}
\label{tab:PerSegEffect}
\end{table}

This observation naturally raises a question:
\textbf{does replacing the redundant or even noisy channels with more effective ones improve the adaptability of foundation models for downstream tasks?} 
To this end, we propose a novel parameter-free fine-tuning paradigm to enhance model adaptability through feature selection, reuse, and enhancement. 
Specifically, we design a channel selection scheme based on the model's output differences to 
separate effective and redundant channels. By selectively replacing redundant channels, our method removes ineffective features and reuses more relevant features to downstream tasks, thereby enhancing task-specific feature representation. 
Figure~\ref{fig:feature comparison} illustrates how our method adapts VFMs to downstream tasks, in contrast to previous fine-tuning that updates parameters.
The contributions of this work are summarized as follows:

\begin{itemize}
    
    \item We 
    identify the feature redundancy in VFMs and 
    propose a parameter-free fine-tuning method for SAM and SAM2. 
    Unlike parameter updating methods, our method focuses on selecting, reusing, and enhancing existing features.

    \item We introduce a channel selection algorithm based on output differences to search for optimal replacement pairs. This process only involves 
    network inference over the search dataset, yielding little computational overhead compared to other fine-tuning approaches.

    \item Extensive experiments on nine diverse datasets demonstrate that the proposed method is simple yet effective. It seamlessly integrates with existing fine-tuning strategies, further boosting the performance of already fine-tuned models.
    
    
\end{itemize}

\section{Related Work}

{\noindent\bf Vision Foundation Models.}
Segment anything model (SAM)~\cite{SAM, SAM2} is one of the most widely used foundational vision models for image segmentation~\cite{SAM}. It generates object masks for images based on user-provided box or point prompts.
SAM's success in segmentation has inspired efforts to extend its application to more complex tasks beyond segmentation, such as tracking and detection~\cite{trackinganything, groundedsam, Inpaintanything}.  
Recently, Meta pushed the boundaries further with the introduction of an even more capable and versatile successor to SAM, known as Segment Anything Model 2 (SAM2), designed specifically for video segmentation~\cite{SAM2}. Similarly, other VFMs such as DINOv2~\cite{dinov2} have demonstrated impressive performance on multiple vision tasks. This paper explores a novel parameter-free fine-tuning method to adapt these foundation models for downstream datasets.

{\noindent\bf  Fine-tuning Strategies.}
Fine-tuning aims to adapt the pre-trained model's weights to specific tasks by selectively updating its parameters.
Existing methods can be categorized into full-parameter and parameter-efficient fine-tuning (PEFT)~\cite{ding2023parameter, li2023loftq, tian2024hydralora}. Full-parameter fine-tuning updates all model parameters, offering strong adaptability but incurring high computational costs. 
As pre-trained models continue to grow in size, PEFT methods, which update only a minimal subset of parameters, have gained importance.
Many PEFT methods are inspired by Adapter~\cite{adapter-icml}, which adapts models by inserting adapter layers into transformer blocks as a few trainable parameters. Since then, various PEFT techniques have been developed, achieving favorable performance, including LoRA~\cite{LoRA,samed}, SVD~\cite{SVDdiff,zelinSVD}, DoRA~\cite{dora}, and Consolidator~\cite{Consolidator}. 
Different from previous  methods that adjust parameters,
In this paper, we introduce a straightforward yet effective fine-tuning method by eliminating feature redundancies.


{\noindent\bf  Channel Operations.} 
Several works have explored channel operations during model training~\cite{zhang2018shufflenet, wang2020deep}. ShuffleNet~\cite{zhang2018shufflenet} introduces a channel shuffle operation, mixing features across groups to ensure cross-channel information fusion. Wang et al.~\cite{wang2020deep} propose a channel-exchanging network, a multimodal fusion framework that dynamically exchanges channels between sub-networks of different modalities.
However, our method significantly differs from these channel shuffle methods in three aspects. (1)  Instead of randomly swapping channels, our method selectively replaces redundant channels with more effective ones. (2) Channel shuffle is used during model training to generate more effective features, whereas our approach reuses the existing features in pre-trained models. (3) Our goal is to boost the adaptability of foundation models for downstream tasks, whereas channel shuffle methods aim to improve information fusion across multiple modalities or within grouped convolutions.


\begin{figure*}[t]
	\begin{minipage}{\linewidth}
	\centering
        \includegraphics[width=1.0\linewidth]{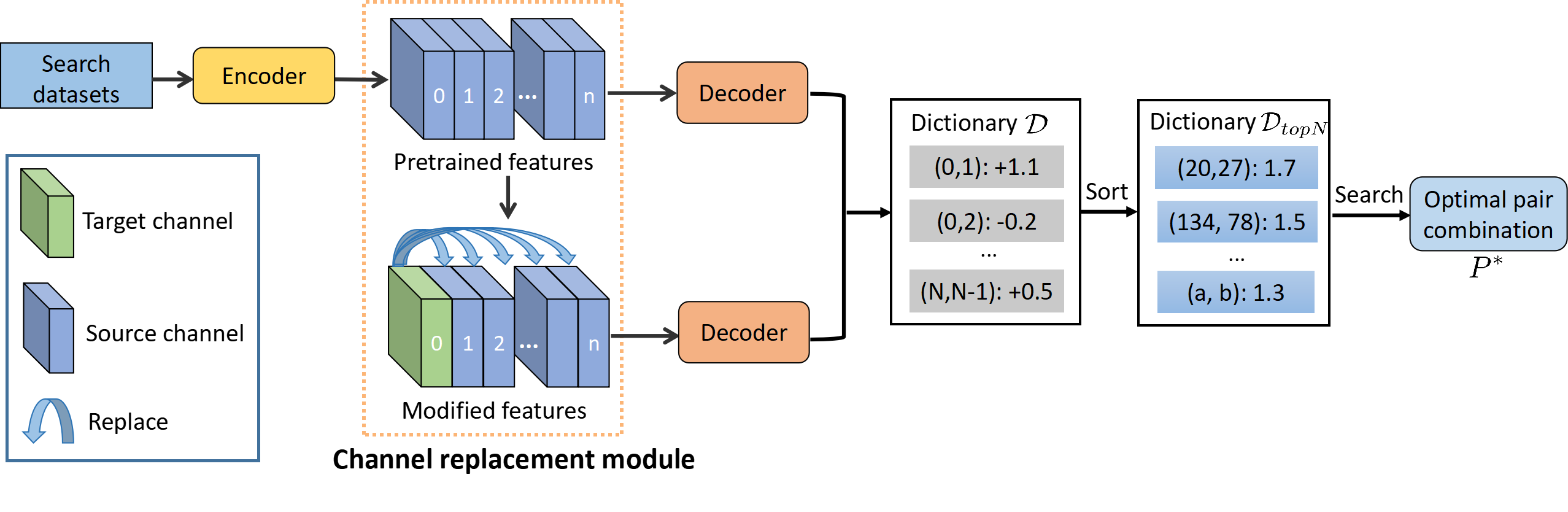}
	\end{minipage}
 \vspace{-0.2in}
	\caption{\textbf{Pipeline of the search process for optimal replacement pair combination.}  
 First, we use a subset of the training dataset, referred to as the ``search dataset,'' which is fed into SAM’s encoder to obtain pretrained features.
 Next, in the channel replacement module, each source channel is sequentially replaced with a target channel, and the modified features are sent to the decoder. By comparing the decoded results from the original and modified features, we construct a dictionary $\mathcal{D}$ that records each replacement pair and its corresponding output difference.  After sorting the top $N$ pairs in $\mathcal{D}$, we further explore the optimal replacement pair combination $P^{*}$ within the subset $\mathcal{D}_{topN}$. By applying $P^{*}$, we replace the redundant channels with more effective ones,  thereby enhancing the performance on downstream tasks.}
	\label{fig:framework}
\end{figure*}

\section{Methodology}

\subsection{Problem Formulation}
Large VFMs like SAM are trained on massive, diverse
training datasets. While this diversity supports generalization,
it also means some channels may carry task-irrelevant
redundant features for specific
downstream tasks. 
\textbf{Our goal is to search for the optimal replacement pairs in which the redundant channels can be replaced by the effective ones, thereby maximizing SAM's performance on downstream tasks.}
We determine a channel to be ``redundant''  if replacing it with another channel improves the model's performance, and hence the replacement channel to be ``effective.'' This implies that the redundant and effective channels always come in pairs. 
Let $P$ represent a single candidate replacement pair combination:
$P = \{(i, j)_1, (i, j)_2, \dots, (i, j)_k \mid i, j, k \in \{1, 2, \ldots, C\} \}$, where $k$ represents the number of replacement pairs, and each $(i, j)$ denotes replacing the feature map of the $i$-th channel with that of the $j$-th channel. $C$ is the total number of channels. The objective function is defined as:
\begin{equation}
P^* = \underset{P}{\arg\max} \ \text{mIoU}(S, P),
\end{equation}
where $P^*$ is the optimal replacement pair combination that maximizes the mean Intersection over Union (mIoU) on the downstream dataset $S$.

However, enumerating all possible pair combinations is computationally infeasible. For instance, when $C = 256$, the number of combinations is $2^{C^{2}}$, which would require an equivalent number of inferences on the training dataset to get an optimal solution.
To address this, we propose several strategies for reducing computational overhead, enabling an efficient search instead of exhaustive enumeration.

\subsection{Channel Selection Algorithm}
{\noindent\bf  Strategies for Reducing Search Overhead.}
To reduce the overhead of searching for the optimal replacement pair combination, we propose the following optimization strategies: (1) \textit{Search based on output difference.} 
First, we calculate the output differences between the original model and the model applying a single replacement pair. We then select the top 
$N$ pairs that yield the highest output improvements. The $N$ pairs form a smaller search space, allowing us to evaluate various combinations of these pairs and find the optimal replacement pair combination. 
This method significantly reduces the number of inferences from $2^{C^{2}}$ to $C^{2}+2^{N}-1$.
(2) \textit{Sample Reduction}. We reduce the original training dataset to a smaller ``search dataset'' containing only 50 images, which helps to reduce the model's inference time.
(3) \textit{Feature Storage}. We pre-store the features of the search dataset generated by the encoder. During each inference, we modify these stored features and pass them to the decoder, avoiding repeated feature extraction by the encoder.
The specific search process is as follows.

\begin{table*}
\centering
\small
\tabcolsep=0.14cm
\begin{tabular}{l|l|c|ccccccccc|l}
\toprule
\textbf{Model} & \textbf{Backbone} & \textbf{Size (M)} & \multicolumn{3}{c|}{\textbf{Natural}} & \multicolumn{3}{c|}{\textbf{Medical}} & \multicolumn{3}{c|}{\textbf{Camou}} & \textbf{Avg + $\Delta$}\\
\cmidrule{4-6} \cmidrule{7-9} \cmidrule{10-12}
& & & COCO & VOC & PerSeg & ISIC16 & BUSI & KVASIR & CAMO & COD10K & CHAME \\
\midrule
\multirow{6}{*}{SAM} & {ViT-B} & {91} & {41.20} &{38.88} & {50.60} & {57.33} & {51.11} & {55.84} & {44.96} &{52.52}   &{49.79}  &{49.14}\\

& $\text{ViT-B}^{*}$ & 91 & \textbf{51.31} & \textbf{56.33} & \textbf{71.68} & \textbf{65.21} & \textbf{58.06} & \textbf{63.08} & \textbf{48.10} & \textbf{55.19} & \textbf{53.78} & \textbf{58.08}\scriptsize{ + \textbf{8.94}}\\ 

& {ViT-L} & {308} & {46.33} & {50.39} & {61.16} & {60.04} & {52.66} & {63.32} & {54.01} & {60.44} & {57.04} & {56.15} \\
& $\text{ViT-L}^{*}$ & 308 & \textbf{{74.82}} &  \textbf{{75.66}} & \textbf{{85.14}} & \textbf{65.77} & \textbf{57.95} & \textbf{66.75} & \textbf{55.27} & \textbf{64.62} & \textbf{62.53} & \textbf{67.61}\scriptsize{ + \textbf{11.46}}  \\

& {ViT-H} & {636} & {46.46} & {49.38} & {62.50} & {61.37} & {50.31} & {62.26} & {52.10} & {61.23} & {54.22} &{55.54} \\ 
& $\text{ViT-H}^{*}$ & 636 & \textbf{54.08} & \textbf{59.18} & \textbf{66.94} & \textbf{63.57} & \textbf{55.33} & \textbf{68.69} & \textbf{54.58} & \textbf{65.82} & \textbf{57.94} & \textbf{60.68}\scriptsize{ + \textbf{5.14}} \\ \midrule
 
\multirow{8}{*}{SAM2} & {Hiera-T} & {39} & {56.53} & {58.16} & {66.67} & {52.18} & {54.82} & {59.32} & {53.98} & {58.50} & {55.47} &{57.29} \\ 
& $\text{Hiera-T}^{*}$ & 39 & \textbf{73.53} & \textbf{76.76} & \textbf{81.77} & \textbf{55.03} & \textbf{57.87} & \textbf{62.08} & \textbf{59.10} & \textbf{64.32} & \textbf{60.18}  & \textbf{65.63}\scriptsize{ + \textbf{8.34}}\\ 

& {Hiera-S} & {46} & {61.72} & {60.88} & {72.45} & {54.38} & {58.20} & {63.43} & {58.30} & {61.38} & {58.59}  &{61.04}\\
& $\text{Hiera-S}^{*}$ & 46 & \textbf{75.41} & \textbf{77.46} & \textbf{84.76} & \textbf{57.78} & \textbf{61.92} & \textbf{66.12} & \textbf{61.91} & \textbf{68.48} & \textbf{64.45}  &\textbf{68.69}\scriptsize{ + \textbf{7.65}}\\ 

& {Hiera-B+} & {81} & {64.92} &{65.00} &{74.23} &{54.55} & {56.46} & {55.83} & {60.38} &62.88  &60.41 &61.62  \\

& $\text{Hiera-B+}^{*}$ & 81 & \textbf{76.18} & \textbf{80.05} & \textbf{75.28} & \textbf{55.86} & \textbf{61.85} & \textbf{59.71} & \textbf{63.70} & \textbf{66.30} & \textbf{63.58}  &\textbf{66.94}\scriptsize{ + \textbf{5.32}}\\ 

& {Hiera-L} & {224} & {71.24} & {70.66} & {77.41} & {58.84} & {66.71} & {60.15} & {67.84} & {69.86} & {67.20} &{67.77} \\
& $\text{Hiera-L}^{*}$ & 224 & \textbf{78.01} & \textbf{83.06} & \textbf{86.27} & \textbf{64.39} & \textbf{69.63} & \textbf{64.49} & \textbf{71.07} & \textbf{73.42} & \textbf{71.49} & \textbf{73.53}\scriptsize{ + \textbf{5.76}}  \\ 
\bottomrule
\end{tabular}
\vspace{-0.05in}
\caption{\textbf{Performance improvement of our fine-tuning for SAM variants on downstream segmentation tasks.} 
``Natural", ``Medical" and ``Camou" represent the natural, medical, and camouflage image datasets, respectively.  ``*" indicates the model of applying our fine-tuning method. ``Size" represents the total parameter of models. ``Avg + $\Delta$" represents the average mIoU along with the improvement over the baseline. Results demonstrate that the proposed method consistently improves segmentation accuracy across different SAM backbones.}
\label{tab:results on SAM-variants}
\end{table*}

{\noindent\bf  Search for Optimal Replacement Pairs.} Given a search dataset $\mathbf{S}$, the feature maps output by SAM's encoder are represented as $X \in \mathbb{R}^{D \times C \times W \times H}$, where $D$ is the total number of search images, $C$ is the total number of channels, and the $W$ and $H$ are the width and height of the feature map, respectively.
Define $\mathcal{P}$ as the set of replacement pairs:
$\mathcal{P} = \{(i, j)\mid i, j \in \{1, 2, \ldots, C\}\}$, where $(i, j)$ represents replacing the feature map of $i$-th channel with that of $j$-th channel. The transformed feature maps can be expressed as:
\begin{equation}
    X'_{d, c, w, h} = X_{d, f_{i \rightarrow j}(c), w, h},
\end{equation}
where $f_{i \rightarrow j}(*)$ is a mapping function that replaces channels according to the replacement pair $(i, j)$.
Then, 
for each replacement pair $(i, j)$,  we calculate the output difference:
\begin{equation}
    \Delta \operatorname{Acc}_{({i} \rightarrow {j})}=\operatorname{D}\left(X'\right)-\operatorname{D}(X),
\end{equation}
where $\operatorname{D}(X)$ and  $\operatorname{D}(X')$ represent the SAM's output using the original feature maps $X$ and the transformed feature maps $X'$, respectively.

To identify the most effective replacement pairs, we compute the accuracy difference for every pair in $\mathcal{P}$, and store them in a dictionary $\mathcal{D}$:
\begin{equation}
    \mathcal{D}=\left\{({i}, {j}): \Delta \operatorname{Acc}_{(i \rightarrow j)}\right\}.
\end{equation}

Next, we select the top $N$ replacement pairs in $\mathcal{D}$ with the highest $\Delta \operatorname{Acc}_{({i} \rightarrow {j})}$ to form a new dictionary:
\begin{equation}
    \mathcal{D}_{topN}=\left\{({i}, {j}): \Delta \operatorname{Acc}_{(i \rightarrow j)} \right\}.
\end{equation}

Finally, we evaluate all possible combinations in $\mathcal{D}_{topN}$ on the downstream dataset to identify the optimal replacement pair combination $P^{*}$, and it can be represented as:
$P^{*} = \{(i, j)_{1}, (i, j)_{2}, ..., (i, j)_{k} \mid (i, j) \in \mathcal{D}_{topN}, k \in \{1, 2, \ldots, N\}\}$. 
Compared to the exhaustive enumeration of all replacement pair combinations, our approach reduces the number of inferences from $2^{C^{2}}$ to $C^{2}+2^{N}-1$, where $C^{2}$ represents the number of elements in $\mathcal{D}$, and $2^{N}-1$ represents the number of non-empty combinations in $\mathcal{D}_{topN}$. 

Figure~\ref{fig:framework} illustrates the complete search process for the optimal replacement pair combination.
According to $P^{*}$, we identify redundant and effective channels in SAM for specific downstream tasks. By selectively replacing redundant channels with more effective ones, we filter out less useful features and reuse more relevant features, thereby enhancing task-specific feature representation.








\begin{table*}
\centering
\small
\tabcolsep=0.16cm
\resizebox{0.99\textwidth}{!}{
\begin{tabular}{l|ccccccccc|l}
\toprule
\textbf{Method}  & \multicolumn{3}{c|}{\textbf{Natural}} & \multicolumn{3}{c|}{\textbf{Medical}} & \multicolumn{3}{c|}{\textbf{Camou}} & \textbf{Avg + $\Delta$} \\
\cmidrule{2-5} \cmidrule{6-8} \cmidrule{9-10}
& COCO & VOC & PerSeg & ISIC16 & BUSI & KVASIR & CAMO & COD10K & CHAME \\
\midrule

\rowcolor{lightgray}
\multicolumn{11}{c}{\textbf{\emph{Partial-parameter Fine-Tuning}}} \\

Decoder  & {69.02} & {75.81} & {90.14} & {80.57} & {78.23} & {75.83} & {63.45} & {65.68} & {63.81}&{73.61} \\

Decoder + Ours  & \textbf{70.26} & \textbf{77.26} & \textbf{90.34} & \textbf{81.41} & \textbf{78.54} & \textbf{77.50} & \textbf{64.70} & \textbf{66.53} & \textbf{65.10} & \textbf{74.62}\scriptsize{ + \textbf{1.01}} \\ 

Encoder  & {51.98} & {72.21} & {89.41} & {85.51}  & {78.61} & {75.94} & {58.64} & {50.65} & {52.03} &{68.33} \\

Encoder + Ours  & \textbf{52.94} & \textbf{72.43} & \textbf{89.56} & \textbf{85.79} & \textbf{78.79} & \textbf{77.16} & \textbf{59.49} & {-} & \textbf{52.77} & \textbf{68.84}\scriptsize{ + \textbf{0.51}}\\ 

MedSAM~\cite{MedSAM}  & {65.19} & {74.28} & {89.94} & {83.78}  & {78.51} & {75.84} & {59.66} & {65.59} & {59.57} &{72.48} \\

MedSAM~\cite{MedSAM} + Ours  & \textbf{67.39} & \textbf{76.38} & \textbf{90.13} & - & \textbf{78.78} & \textbf{77.30} & \textbf{60.46} & \textbf{67.60} & \textbf{63.59} & \textbf{73.93}\scriptsize{ + \textbf{1.45}}\\ 

\rowcolor{lightgray}
\multicolumn{11}{c}{\textbf{\emph{Parameter-efficient Fine-Tuning}}}
\\

$\text{SAMed}$~\cite{samed}  & {66.86} & {77.38} & {89.81} & {85.35} & {81.01} & {83.39} & {74.53} & {75.47} & {73.25} &{78.56}\\

$\text{SAMed}$~\cite{samed} + Ours  & \textbf{70.29}  & \textbf{78.85} & \textbf{90.19} & \textbf{86.27} & \textbf{82.26} & \textbf{83.86} & \textbf{75.17} & \textbf{76.50} & \textbf{74.12} & \textbf{79.72}\scriptsize{ + \textbf{1.16}}\\

SAM-COBOT~\cite{SAM-COBOT}   & {68.96} & {82.30} & {91.20} & {86.31} & {78.44} & {83.59} & {71.41} & {73.88} & {72.47} &{78.73}\\

SAM-COBOT~\cite{SAM-COBOT} + Ours  & \textbf{69.41} & \textbf{82.87} & \textbf{92.09} & \textbf{86.57} & \textbf{79.54} & \textbf{84.07} & \textbf{71.94} & \textbf{74.50} & \textbf{72.93} & \textbf{79.32}\scriptsize{ + \textbf{0.59}}\\

SAM-Adapter~\cite{Sam-adapter} & {65.76} & {74.24} & {80.48} & {83.59} & {72.83} & {70.77} & {67.79} & {71.20} & {69.39} & {72.89}\\

SAM-Adapter~\cite{Sam-adapter} + Ours  & \textbf{67.28} & \textbf{75.85} & \textbf{82.11} & \textbf{83.91} & \textbf{73.10} & \textbf{71.69} & \textbf{68.05} & \textbf{72.44} &  \textbf{69.80} & \textbf{73.80}\scriptsize{ + \textbf{0.91}}\\

SAM-PARSER~\cite{zelinSVD}  &63.14   &67.24   &55.82   &71.62  &57.66   &62.69   &53.30   &60.99 &56.20  & {60.96}\\

SAM-PARSER~\cite{zelinSVD} + Ours  & \textbf{68.84} & \textbf{70.63} & \textbf{70.29} & \textbf{72.88} & \textbf{62.42} & \textbf{67.87} & \textbf{55.85} & \textbf{61.40} & \textbf{58.31} & \textbf{65.39}\scriptsize{ + \textbf{4.43}}\\

DoRA~\cite{dora}  & 67.33 & 77.95 & 91.24 & 85.68 & 81.53 & 83.83 & 75.05 & 75.80 & 73.71 & 79.12\\

DoRA~\cite{dora} + Ours  & \textbf{69.16} & \textbf{80.19} & \textbf{91.63} & \textbf{85.71} & \textbf{82.65} & \textbf{84.30} & \textbf{75.54} & \textbf{75.94} & \textbf{74.17} & \textbf{79.92}\scriptsize{ + \textbf{0.80}}\\

\bottomrule
\end{tabular}
}
\vspace{-0.05in}
\caption{\textbf{Performance improvement for fine-tuned SAM on downstream segmentation tasks.} All results are based on the ViT-Base version of SAM. ``Decoder'': fine-tuning only decoder. ``Encoder'': fine-tuning only encoder.   ```Avg + $\Delta$": the average mIoU along with the improvement over the baseline. ``-'': without any improvement.  It shows that our method can serve as a plug-and-play module that further enhances the performance of fine-tuned models.}
\label{tab:performance comparison of different fine-tuning}
\end{table*}
\section{Experiments}
\subsection{Experimental Setup}
{\noindent\bf  Datasets and Metrics.} We evaluate our method across 9 datasets, covering different tasks of natural image segmentation  (COCO~\cite{COCOdataset}, VOC~\cite{VOCdataset} and PerSeg~\cite{PerSAM}),  medical image segmentation (ISIC~\cite{ISIC}, BUSI~\cite{BUSI} and KVASIR~\cite{jha2020kvasir}) and camouflage detection (CAMO~\cite{CAMO}, COD10K~\cite{COD} and CHAME~\cite{Chameleon}).  Each image is paired with a ground truth segmentation mask of a single object, along with corresponding prompts.
Natural image datasets are used to assess SAM’s performance under in-domain scenarios, while medical and camouflage datasets are used to evaluate SAM's performance under out-of-domain scenarios. All tasks use mean Intersection over Union (mIoU) as evaluation metrics.

{\noindent\bf  Baselines.} To validate the effectiveness of our method, we choose SAM~\cite{SAM} and SAM 2~\cite{SAM2} as baseline models with different backbones for the segmentation task. The backbones of SAM include ViT-B, ViT-L, and ViT-H, whereas SAM 2 includes Hiera-T, Hiera-S, Hiera-B+, and Hiera-L. We also choose Dinov2~\cite{dinov2} as baseline model for depth estimation and image classification tasks. To validate that our method can further improve the performance of already fine-tuned models, we introduce various parameter-efficient fine-tuning schemes including SAMed~\cite{samed}, SAM-Adapter~\cite{Sam-adapter}, SAM-PARSER~\cite{zelinSVD}, SAM-COBOT~\cite{SAM-COBOT}, DORA~\cite{dora} and partial-parameter fine-tuning methods, including MedSAM~\cite{MedSAM}, Encoder-only Tuning, and Decoder-only Tuning. 


{\noindent\bf  Implementation Details.} For comparison to other fine-tuning, we employ the ViT-Base version of SAM~\cite{SAM} as our backbones, integrating point prompts for the prompt encoder input. For loss functions, we use the same combination of Dice loss and CE loss as in~\cite{samed, SAM-COBOT}. For the search dataset, we randomly select 50 image samples from the training dataset. $N$ in $\mathcal{D}_{topN}$ is set to 10. For the training epoch, we set the total number of epochs to 25. Our training employs the Adam optimizer~\cite{adam}. The initial learning rate is set to \(1.0 \times 10^{-4}\), and the weight decay is \(5 \times 10^{-5}\) with one image per mini-batch.  All experiments are conducted on a computer equipped with four NVIDIA RTX 4090 GPUs. Please refer to the Appendix for more details.

\subsection{Main Results}

{\noindent\bf Benchmark Results.} Table~\ref{tab:results on SAM-variants} quantitatively compares various SAM-variants on downstream segmentation tasks across natural, medical, and camouflage image datasets. The results validate the effectiveness of our proposed method in consistently enhancing the segmentation capability of SAM and SAM 2 across diverse datasets, including COCO~\cite{COCOdataset}, VOC~\cite{VOCdataset}, PerSeg~\cite{PerSAM}, ISIC~\cite{ISIC}, BUSI~\cite{BUSI}, KVASIR~\cite{jha2020kvasir}, CAMO~\cite{CAMO}, COD10K~\cite{COD} and CHAME~\cite{Chameleon}. Specifically, we observe that our method is effective across different backbones. It enhances the average mIoU of ViT-B, Vit-L, and ViT-H versions of SAM by 8.94, 11.46, and 5.14, respectively. Similarly, 
it enhances the average mIoU of Hiera-T, Hiera-S, Hiera-B+, and Hiera-L versions of SAM 2 by 8.34, 7.65, 5.32, and 5.76, respectively. 
Moreover, larger improvements are observed on natural image datasets. For example, on the ISIC16 medical dataset, the ViT-B version of SAM improves from 57.33 to 65.21 (+7.88), while on the VOC natural image dataset, it improves from 38.88 to 56.33 (+17.45). 
This suggests that our fine-tuning method is more effective on in-domain datasets (e.g.,, VOC) than out-of-domain datasets (e.g.,, ISIC16).


\begin{figure*}[t]
	\begin{minipage}{\linewidth}
	\centering
        \includegraphics[width=0.98\linewidth]{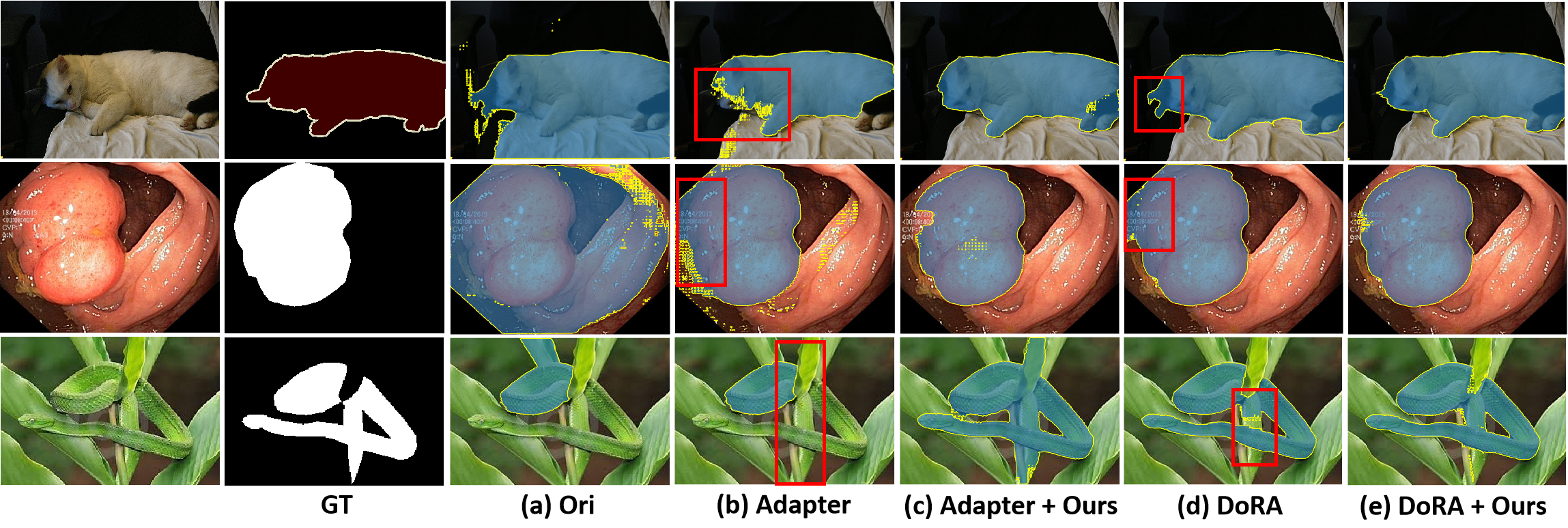}
	\end{minipage}
        \vspace{-0.1in}
	\caption{\textbf{Qualitative comparison of various fine-tuning methods for SAM across natural, medical, and camouflage scenarios.} Columns from left to right show the original input image (Input), the ground-truth segmentation mask (GT), the segmentation results from the original SAM (Base), and the results before and after applying our method to various fine-tuned methods (i.e., Adapter, DoRA). The results demonstrate that our method effectively enhances the performance of already fine-tuned models,  producing more refined predictions that are closer to the ground truths, as highlighted in  \textcolor{red}{Red boxes}. Refer to the Appendix for more visualizations. 
 }
	\label{fig:Qualitative results}
 \vspace{-0.10in}
\end{figure*}

{\noindent\bf Enhancing fine-tuned Models.} Table~\ref{tab:performance comparison of different fine-tuning} compares the performance of various fine-tuning strategies for SAM, including  Decoder-only Tuning, Encoder-only Tuning, MedSAM~\cite{MedSAM}, SAMed~\cite{samed}, SAM-COBOT~\cite{SAM-COBOT}, SAM-Adapter~\cite{Sam-adapter}, SAM-PARSER~\cite{zelinSVD}, and DoRA~\cite{dora}, across nine datasets spanning three computer vision domains. In Table~\ref{tab:performance comparison of different fine-tuning}, we present both the original mIoU and the mIoU after applying our method to these fine-tuned models. The results demonstrate that our method effectively enhances these fine-tuned models, leading to consistent performance improvements across various tasks. For example, our method boosts SAMed, SAM-Adapter, SAM-PARSER and DoRA by 1.16, 0.91, 4.43 and 0.80, respectively, as well as Decoder-tuning and Encoder-tuning by 1.01 and 0.51. These findings suggest that feature redundancy persists in the models even after fine-tuning. By effectively identifying and removing this redundancy, our approach enhances the models' ability to better adapt to downstream tasks.

{\noindent\bf Qualitative Results.} 
Figure~\ref{fig:Qualitative results} presents segmentation examples that demonstrate the effectiveness of our method compared to leading fine-tuning techniques, including DoRA~\cite{dora}, and SAM-Adapter~\cite{adapter-icml}, across natural, medical and camouflage scenarios. In Figure~\ref{fig:Qualitative results}, our method is compatible with existing fine-tuning across various scenarios, achieving a more precise segmentation. For instance, details such as the cat’s ear, the polyp’s contour, and the snake’s body align more closely with the ground truths.

\begin{figure}[t]
	\begin{minipage}{\linewidth}
	\centering
\includegraphics[width=1.0\linewidth]{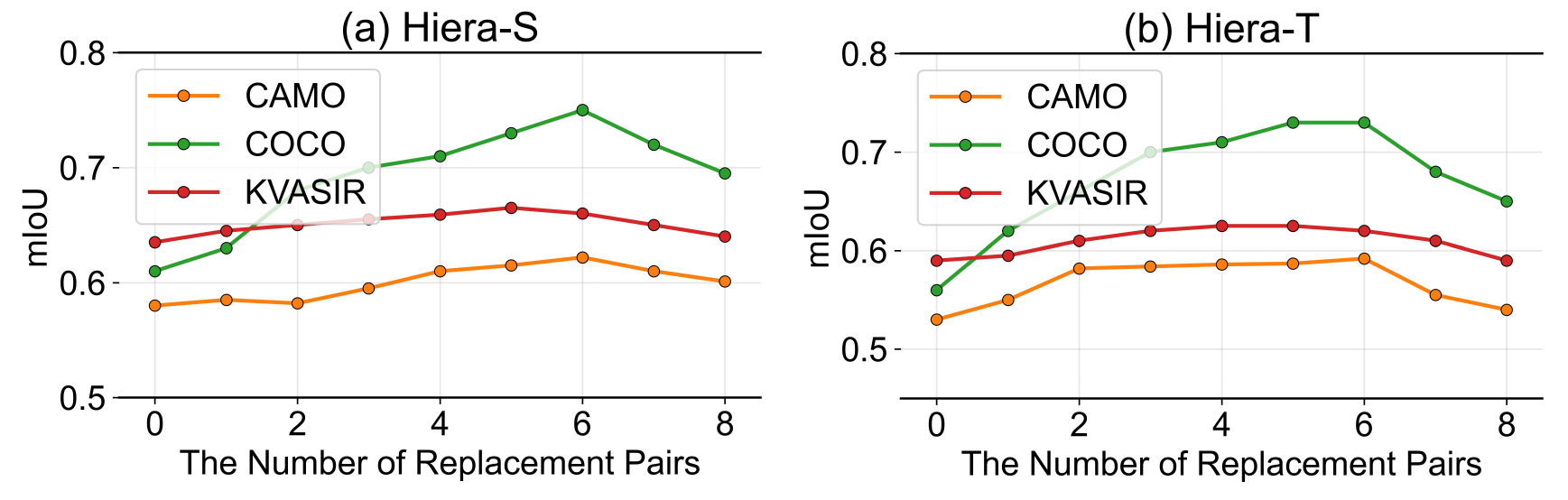}
	\end{minipage}	
    \vspace{-0.1in}
    \caption{\textbf{Performance Comparison of varying number of replacement pairs.} This suggests that combining different replacement pairs can further improve segmentation performance.}
    \label{fig:varying number of replacement pairs}
\end{figure}

\subsection{Ablation Studies}

\begin{figure*}[t]
    \centering
    \includegraphics[width=0.99\linewidth]{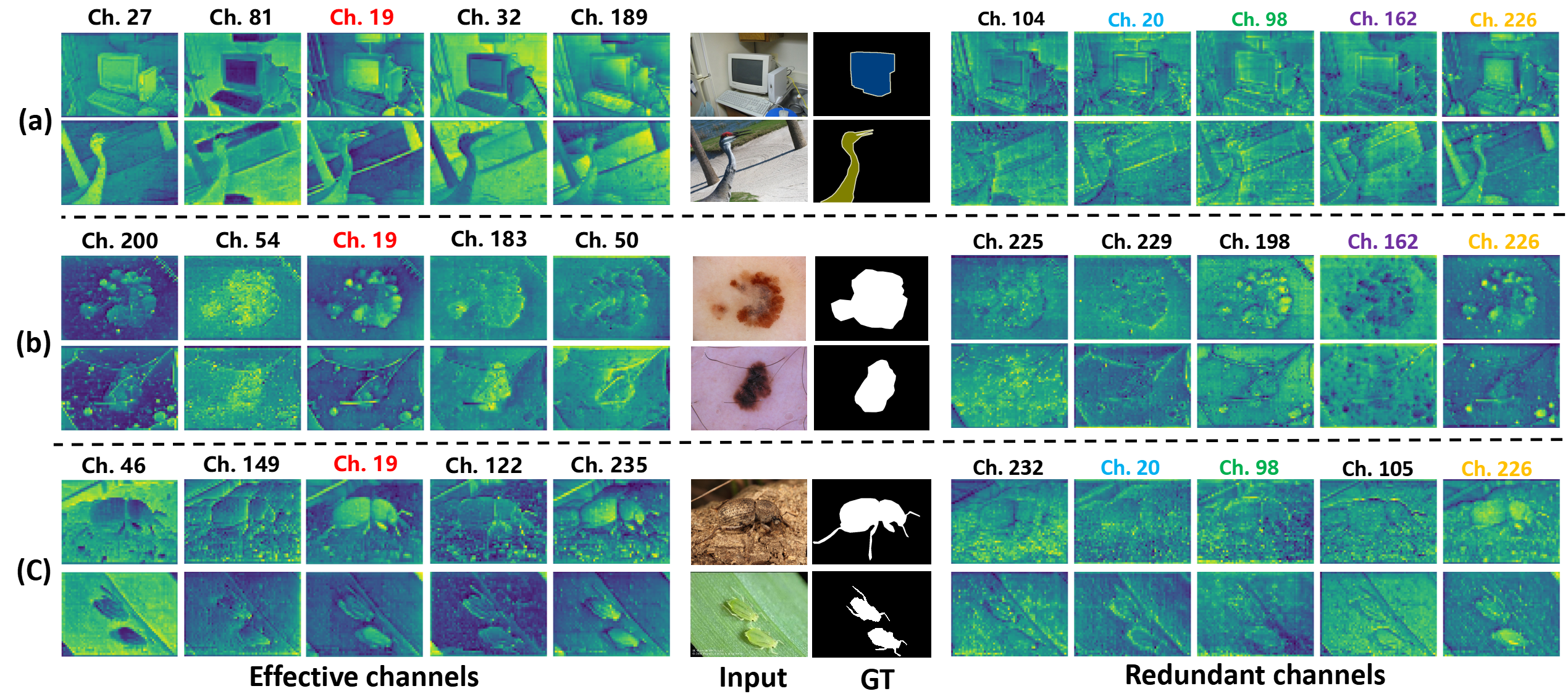}
    \vspace{-0.1in}
    \caption{\textbf{Comparison of effective and redundant channels in optimal replacement pairs for downstream tasks.} ``Effective channels'' are used
to replace the ``Redundant channels'' to improve task-specific performance. The colors in the feature map from \textcolor{darkgreen}{green} to \textcolor{goldenyellow}{yellow} represent the response intensity from weak to strong. 
    (a)-(c) showcase natural, medical and camouflage scenarios, respectively,  showing effective channel features have more discernible structures, edges and textures compared to redundant channel features. }
    \label{fig:Visualization of redundant and effective features}
\end{figure*}

\begin{table}[t]
\centering
\small
\resizebox{0.38\textwidth}{!}{
\begin{tabular}{c|ccc}
\toprule
Method  & GPU (GB) & Params.(K) \\ \midrule
Encoder-only   &34.6  &89670  \\
Decoder-only   &13.7  &4057  \\
MedSAM~\cite{MedSAM}   &34.7  &93735  \\
SAM-Adapter~\cite{Sam-adapter}   &28.5  &3550  \\
SAMed~\cite{samed} &28.9  &147  \\
SAM-COBOT~\cite{SAM-COBOT} &29.1 &148 \\
SAM-PARSER~\cite{zelinSVD} &15.9 &0.5 \\
Ours   &11.1  &0  \\ 
\bottomrule
\end{tabular}
}
\vspace{-0.1in}
\caption{\textbf{Computational overhead for different fine-tuning strategies.} Experiments are performed on the ViT-base version of SAM with input images at a resolution of 1024x1024 and a batch size of 4.  It shows that our method requires less GPU usage and is entirely parameter-free compared to other fine-tuning.}
\label{tab:Computational overhead}
\vspace{-0.15in}
\end{table}

{\noindent\bf Varying of Number of Replacement Pairs.}
To examine the relationship between the number of replacement pairs and performance, we evaluate different replacement pair combinations across various datasets (e.g., CAMO, COCO, KVASIR) using SAM's Hiera-S and Hiera-T backbones, as shown in Figure~\ref{fig:varying number of replacement pairs}.
The results indicate that increasing the number of replacement pairs leads to improved performance. For example, on the COCO dataset (the green line), performance peaks when using six replacement pairs. This suggests that combining different replacement pairs effectively reduces redundancy in SAM.


{\noindent\bf Computational Overhead.} Table~\ref{tab:Computational overhead} presents the computational overhead comparison of different fine-tuning methods. The results demonstrate that our method significantly reduces GPU memory usage compared to other fine-tuning approaches. While some PEFT methods (i.e., SAMed and SAM-PARSER) update only a small subset of parameters, they still require retaining the original computational graph for backpropagation, leading to high GPU memory consumption. In contrast, our method is entirely parameter-free, relying only on model inference, which minimizes memory usage and eliminates the need for additional parameters for adaptation.

{\noindent\bf Features of Optimal Replacement Pairs.} 
To explore which feature types are redundant or effective, we compare channels in optimal replacement pairs on natural, medical, and camouflage scenarios. As shown in Figure~\ref{fig:Visualization of redundant and effective features}, the features of effective channels have more discernible structures, well-defined edges, and textures, which are likely beneficial for segmentation tasks. In contrast, the features of redundant channels are generally more blurred and contain less structural information. They appear noisy and lack the clarity for accurate segmentation. Moreover, some certain channels exhibit consistent effectiveness or redundancy across various datasets. For example, Channel 19 shows consistent effectiveness, while Channels 20, 98, 162, and 226 show consistent redundancy across different scenarios. It indicates that some channels capture features with cross-domain generalizability.

{\noindent\bf Extension to Depth Estimation and Image classification.}
To validate the effectiveness of our method beyond segmentation, we evaluate its performance on additional vision tasks, including image classification and depth estimation. Table~\ref{tab:depth estimation and classification} presents the depth estimation results on the NYUv2 dataset~\cite{NYUv2} and the image classification results on the CIFAR dataset~\cite{Cifar}. Our method improves the accuracy of the ViT-Small and ViT-Base versions of DINOv2 from 80.41 to 80.81 and from 88.08 to 88.49, respectively. Similarly, it enhances the MSE, AbsRel, and $\delta_{1}$ metrics from 0.210, 0.110, and 0.900 to 0.193, 0.095, and 0.916, respectively. These results demonstrate the effectiveness of our approach across different vision tasks (See Appendix for the visualization).

\begin{table}[t]
\centering
\small
\resizebox{0.43\textwidth}{!}{
\begin{tabular}{c|l|c|c|c|c}
\toprule
Model & Backbone & \multicolumn{3}{c|}{NYUv2} & \multicolumn{1}{c}{CIFAR} \\
\cmidrule{3-6}
 &  & MSE $\downarrow$ & AbsRel $\downarrow$ & $\delta_{1}$ $\uparrow$  & Acc.$\uparrow$\\
\midrule
\multirow{2}{*}{\centering Dinov2} 
       & ViT-S &0.225&0.126&0.893&80.41\\
       & $\text{ViT-S}^{*}$ &\textbf{0.209}&\textbf{0.112}&\textbf{0.907} &\textbf{80.81}\\
       & ViT-B  & 0.210 & 0.110  & 0.900 & 88.08  \\
       & $\text{ViT-B}^{*}$ & \textbf{0.193} & \textbf{0.095}  & \textbf{0.916} & \textbf{88.49} \\
\bottomrule
\end{tabular}
}
\vspace{-0.1in}
\caption{\textbf{Effectiveness evaluation of Depth Estimation and Image Classification.} This demonstrates the generalizability of our fine-tuning approach across depth estimation and classification tasks.}
\label{tab:depth estimation and classification}
\end{table}



\section{Discussion}
Our method achieves performance gains without updating any parameters by exploiting a key observation: not all features in vision foundation models are equally useful for downstream tasks. Large VFMs like SAM are trained on massive, diverse
training datasets and are thus equipped with rich but overcomplete
representations. While this diversity supports generalization,
it also means many channels are not tailored for specific
downstream tasks. Therefore, some channels may carry task-irrelevant
or even distracting information—what we define as redundant. \textbf{\textit{From an intuitive perspective},} we use the replacing strategy to reconfigure the existing representation space. By replacing these redundant channels with effective ones already existing in the model, we enhance task-specific feature representation. This can be analogized to team optimization: if some team members are idle or misaligned with the current project goals, reallocating active contributors to fill their roles enhances team efficiency—without hiring anyone new. \textbf{\textit{From the theoretical perspective,}} our method is grounded in the Information Bottleneck (IB) theory~\cite{tishby2000information}, which seeks to maximize the mutual information with the output labels $I(X; Y)$ while minimizing that with the input $I(X; I)$. This reflects a trade-off between preserving task-relevant information and discarding redundant input features.
Our approach aligns with this principle by identifying channels with low task relevance—i.e., low $I(X; Y)$ or high $I(X; I)$—and replacing them with more informative ones, thereby enhancing the downstream representation.






\section{Conclusion}
In this paper, we propose a parameter-free fine-tuning method for adapting vision foundation models (VFMs) to downstream tasks. Unlike previous fine-tuning methods that update parameters, our approach introduces a new perspective by eliminating feature redundancies within VFMs. Experiments on both in-domain and out-of-domain datasets demonstrate that our method is simple yet effective, and it is also compatible with existing fine-tuning strategies to achieve additional performance gains.
Since our method operates solely at the inference stage, it maintains a lower computational overhead.
Future work could explore accelerating the search for optimal replacement pairs using traditional optimization techniques, such as particle swarm optimization. 
Additionally, given its strong generalization across multiple backbones and downstream tasks, we encourage researchers to apply our method to a broader range of models and vision tasks.

\bibliography{aaai2026}

@String(AAAI = {AAAI})

@article{dinov2,
  title={Dinov2: Learning robust visual features without supervision},
  author={Oquab, Maxime and Darcet, Timoth{\'e}e and Moutakanni, Th{\'e}o and Vo, Huy and Szafraniec, Marc and Khalidov, Vasil and Fernandez, Pierre and Haziza, Daniel and Massa, Francisco and El-Nouby, Alaaeldin and others},
  journal={arXiv:2304.07193},
  year={2023}
}

@article{SAM,
  title={Segment anything},
  author={Kirillov, Alexander and Mintun, Eric and Ravi, Nikhila and Mao, Hanzi and Rolland, Chloe and Gustafson, Laura and Xiao, Tete and Whitehead, Spencer and Berg, Alexander C and Lo, Wan-Yen and others},
  journal={arXiv:2304.02643},
  year={2023}
}

@article{samed,
  title={Customized segment anything model for medical image segmentation},
  author={Zhang, Kaidong and Liu, Dong},
  journal={arXiv:2304.13785},
  year={2023}
}

@inproceedings{Sam-adapter,
  title={Sam-adapter: Adapting segment anything in underperformed scenes},
  author={Chen, Tianrun and Zhu, Lanyun and Deng, Chaotao and Cao, Runlong and Wang, Yan and Zhang, Shangzhan and Li, Zejian and Sun, Lingyun and Zang, Ying and Mao, Papa},
  booktitle={Proceedings of the IEEE/CVF International Conference on Computer Vision},
  pages={3367--3375},
  year={2023}
}

@article{SAM-adapter-Camouflage-Shadow-more,
  title={SAM Fails to Segment Anything?--SAM-Adapter: Adapting SAM in Underperformed Scenes: Camouflage, Shadow, and More},
  author={Chen, Tianrun and Zhu, Lanyun and Ding, Chaotao and Cao, Runlong and Zhang, Shangzhan and Wang, Yan and Li, Zejian and Sun, Lingyun and Mao, Papa and Zang, Ying},
  journal={arXiv:2304.09148},
  year={2023}
}

@article{VOCdataset,
  title={The pascal visual object classes (VOC) challenge},
  author={Everingham, Mark and Van Gool, Luc and Williams, Christopher KI and Winn, John and Zisserman, Andrew},
  journal={International journal of computer vision},
  year={2010}
}

@inproceedings{COCOdataset,
  title={Microsoft coco: Common objects in context},
  author={Lin, Tsung-Yi and Maire, Michael and Belongie, Serge and Hays, James and Perona, Pietro and Ramanan, Deva and Doll{\'a}r, Piotr and Zitnick, C Lawrence},
  booktitle={European Conference on Computer Vision},
  year={2014}
}

@article{LoRA,
  title={Lora: Low-rank adaptation of large language models},
  author={Hu, Edward J and Shen, Yelong and Wallis, Phillip and Allen-Zhu, Zeyuan and Li, Yuanzhi and Wang, Shean and Wang, Lu and Chen, Weizhu},
  journal={arXiv:2106.09685},
  year={2021}
}

@article{adam,
  title={Adam: A method for stochastic optimization},
  author={Kingma, Diederik P and Ba, Jimmy},
  journal={arXiv:1412.6980},
  year={2014}
}

@inproceedings{adapter-icml,
  title={Parameter-efficient transfer learning for NLP},
  author={Houlsby, Neil and Giurgiu, Andrei and Jastrzebski, Stanislaw and Morrone, Bruna and De Laroussilhe, Quentin and Gesmundo, Andrea and Attariyan, Mona and Gelly, Sylvain},
  booktitle={International Conference on Machine Learning},
  year={2019},
}

@inproceedings{SAM-COBOT,
  title={Parameter efficient fine-tuning via cross block orchestration for segment anything model},
  author={Peng, Zelin and Xu, Zhengqin and Zeng, Zhilin and Xie, Lingxi and Tian, Qi and Shen, Wei},
  booktitle={Proceedings of the IEEE/CVF Conference on Computer Vision and Pattern Recognition},
  year={2024}
}

@article{SAM2,
  title={Sam 2: Segment anything in images and videos},
  author={Ravi, Nikhila and Gabeur, Valentin and Hu, Yuan-Ting and Hu, Ronghang and Ryali, Chaitanya and Ma, Tengyu and Khedr, Haitham and R{\"a}dle, Roman and Rolland, Chloe and Gustafson, Laura and others},
  journal={arXiv:2408.00714},
  year={2024}
}

@article{MedSAM,
  title={Segment anything in medical images},
  author={Ma, Jun and He, Yuting and Li, Feifei and Han, Lin and You, Chenyu and Wang, Bo},
  journal={Nature Communications},
  year={2024},
  publisher={Nature Publishing Group UK London}
}

@article{trackinganything,
  title={Track anything: Segment anything meets videos},
  author={Yang, Jinyu and Gao, Mingqi and Li, Zhe and Gao, Shang and Wang, Fangjing and Zheng, Feng},
  journal={arXiv:2304.11968},
  year={2023}
}

@article{Inpaintanything,
  title={Inpaint anything: Segment anything meets image inpainting},
  author={Yu, Tao and Feng, Runseng and Feng, Ruoyu and Liu, Jinming and Jin, Xin and Zeng, Wenjun and Chen, Zhibo},
  journal={arXiv:2304.06790},
  year={2023}
}

@article{groundedsam,
  title={Grounded sam: Assembling open-world models for diverse visual tasks},
  author={Ren, Tianhe and Liu, Shilong and Zeng, Ailing and Lin, Jing and Li, Kunchang and Cao, He and Chen, Jiayu and Huang, Xinyu and Chen, Yukang and Yan, Feng and others},
  journal={arXiv:2401.14159},
  year={2024}
}

@inproceedings{CLIP,
  title={Learning transferable visual models from natural language supervision},
  author={Radford, Alec and Kim, Jong Wook and Hallacy, Chris and Ramesh, Aditya and Goh, Gabriel and Agarwal, Sandhini and Sastry, Girish and Askell, Amanda and Mishkin, Pamela and Clark, Jack and others},
  booktitle={International conference on machine learning},
  year={2021},
}

@inproceedings{SVDdiff,
  title={Svdiff: Compact parameter space for diffusion fine-tuning},
  author={Han, Ligong and Li, Yinxiao and Zhang, Han and Milanfar, Peyman and Metaxas, Dimitris and Yang, Feng},
  booktitle={Proceedings of the IEEE/CVF International Conference on Computer Vision},
  year={2023}
}

@article{PerSAM,
  title={Personalize segment anything model with one shot},
  author={Zhang, Renrui and Jiang, Zhengkai and Guo, Ziyu and Yan, Shilin and Pan, Junting and Ma, Xianzheng and Dong, Hao and Gao, Peng and Li, Hongsheng},
  journal={arXiv:2305.03048},
  year={2023}
}

@inproceedings{ISIC,
  title={Skin lesion analysis toward melanoma detection},
  author={Codella, Noel CF and Gutman, David and Celebi, M Emre and Helba, Brian and Marchetti, Michael A and Dusza, Stephen W and Kalloo, Aadi and Liopyris, Konstantinos and Mishra, Nabin and Kittler, Harald and others},
  booktitle={IEEE 15th international symposium on biomedical imaging},
  year={2018},
}

@article{BUSI,
  title={Dataset of breast ultrasound images},
  author={Al-Dhabyani, Walid and Gomaa, Mohammed and Khaled, Hussien and Fahmy, Aly},
  journal={Data in brief},
  year={2020},
  publisher={Elsevier}
}

@inproceedings{jha2020kvasir,
  title={Kvasir-seg: A segmented polyp dataset},
  author={Jha, Debesh and Smedsrud, Pia H and Riegler, Michael A and Halvorsen, P{\aa}l and De Lange, Thomas and Johansen, Dag and Johansen, H{\aa}vard D},
  booktitle={MultiMedia modeling: 26th international conference},
  year={2020},
}

@article{CAMO,
  Title          = {Anabranch Network for Camouflaged Object Segmentation},
  Author         = {Trung-Nghia Le and Tam V. Nguyen and Zhongliang Nie and Minh-Triet Tran and Akihiro Sugimoto},
  Journal        = {Journal of Computer Vision and Image Understanding},
  Year           = {2019},
}

@article{COD,
               author={Fan, Deng-Ping and Ji, Ge-Peng and Cheng, Ming-Ming and Shao, Ling},
               title={Concealed object detection},
               journal={IEEE Transactions on Pattern Analysis and Machine Intelligence},
               year={2022}
}

@article{Chameleon,
  title={Animal camouflage analysis: Chameleon database},
  author={Skurowski, Przemys{\l}aw and Abdulameer, Hassan and B{\l}aszczyk, J and Depta, Tomasz and Kornacki, Adam and Kozie{\l}, P},
  journal={Unpublished manuscript},
  year={2018}
}

@article{ding2023parameter,
  title={Parameter-efficient fine-tuning of large-scale pre-trained language models},
  author={Ding, Ning and Qin, Yujia and Yang, Guang and Wei, Fuchao and Yang, Zonghan and Su, Yusheng and Hu, Shengding and Chen, Yulin and Chan, Chi-Min and Chen, Weize and others},
  journal={Nature Machine Intelligence},
  year={2023},
  publisher={Nature Publishing Group UK London}
}

@inproceedings{zhang2018shufflenet,
  title={Shufflenet: An extremely efficient convolutional neural network for mobile devices},
  author={Zhang, Xiangyu and Zhou, Xinyu and Lin, Mengxiao and Sun, Jian},
  booktitle={Proceedings of the IEEE conference on computer vision and pattern recognition},
  year={2018}
}

@article{wang2020deep,
  title={Deep multimodal fusion by channel exchanging},
  author={Wang, Yikai and Huang, Wenbing and Sun, Fuchun and Xu, Tingyang and Rong, Yu and Huang, Junzhou},
  journal={Advances in neural information processing systems},
  year={2020}
}

@inproceedings{zelinSVD,
  title={Sam-parser: Fine-tuning sam efficiently by parameter space reconstruction},
  author={Peng, Zelin and Xu, Zhengqin and Zeng, Zhilin and Yang, Xiaokang and Shen, Wei},
  booktitle={Proceedings of the AAAI Conference on Artificial Intelligence},
  year={2024}
}

@inproceedings{li2023scconv,
  title={Scconv: Spatial and channel reconstruction convolution for feature redundancy},
  author={Li, Jiafeng and Wen, Ying and He, Lianghua},
  booktitle={Proceedings of the IEEE/CVF Conference on Computer Vision and Pattern Recognition},
  year={2023}
}

@inproceedings{dalvi2020analyzing,
  title={Analyzing redundancy in pretrained transformer models},
  author={Dalvi, Fahim and Sajjad, Hassan and Durrani, Nadir and Belinkov, Yonatan},
  booktitle={Proceedings of the 2020 conference on empirical methods in natural language processing},
  year={2020}
}

@inproceedings{Consolidator,
  title={Consolidator: Mergeable Adapter with Grouped Connections for Visual Adaptation},
  author={{Tianxiang Hao} and {Hui Chen} and {Yuchen Guo} and {Guiguang Ding}},
  booktitle={The Eleventh International Conference on Learning Representations},
  year={2023}
}

@inproceedings{xiong2025pyra,
  title={Pyra: Parallel yielding re-activation for training-inference efficient task adaptation},
  author={Xiong, Yizhe and Chen, Hui and Hao, Tianxiang and Lin, Zijia and Han, Jungong and Zhang, Yuesong and Wang, Guoxin and Bao, Yongjun and Ding, Guiguang},
  booktitle={European Conference on Computer Vision},
  year={2024},
}

@inproceedings{NYUv2,
  title={Indoor segmentation and support inference from rgbd images},
  author={Silberman, Nathan and Hoiem, Derek and Kohli, Pushmeet and Fergus, Rob},
  booktitle={12th European Conference on Computer Vision},
  year={2012},
}

@article{Cifar,
  title={Learning multiple layers of features from tiny images},
  author={Krizhevsky, Alex and Hinton, Geoffrey and others},
  year={2009},
}

@inproceedings{dora,
  title={Dora: Weight-decomposed low-rank adaptation},
  author={Liu, Shih-Yang and Wang, Chien-Yi and Yin, Hongxu and Molchanov, Pavlo and Wang, Yu-Chiang Frank and Cheng, Kwang-Ting and Chen, Min-Hung},
  booktitle={Forty-first International Conference on Machine Learning},
  year={2024}
}

@article{tishby2000information,
  title={The Information Bottleneck Method},
  author={Tishby, Naftali and Pereira, Fernando C and Bialek, William},
  journal={arXiv preprint physics/0004057},
  year={2000}
}

@article{li2023loftq,
  title={Loftq: Lora-fine-tuning-aware quantization for large language models},
  author={Li, Yixiao and Yu, Yifan and Liang, Chen and He, Pengcheng and Karampatziakis, Nikos and Chen, Weizhu and Zhao, Tuo},
  journal={arXiv:2310.08659},
  year={2023}
}

@article{tian2024hydralora,
  title={Hydralora: An asymmetric lora architecture for efficient fine-tuning},
  author={Tian, Chunlin and Shi, Zhan and Guo, Zhijiang and Li, Li and Xu, Cheng-Zhong},
  journal={Advances in Neural Information Processing Systems},
  year={2024}
}

\clearpage	
\newpage

\end{document}